\documentclass[conference]{IEEEtran}
\IEEEoverridecommandlockouts

\usepackage{cite}
\usepackage{amsmath,amssymb,amsfonts}
\usepackage{graphicx}
\usepackage{textcomp}
\usepackage{xcolor}
\usepackage[linesnumbered,ruled]{algorithm2e}
\usepackage{tabu,booktabs}
\usepackage{hyperref}
\hypersetup{
     colorlinks=true,
     linkcolor=red,
     filecolor=blue,
     citecolor=blue,      
     urlcolor=magenta,
     }

\usepackage{subcaption}

\def\BibTeX{{\rm B\kern-.05em{\sc i\kern-.025em b}\kern-.08em
    T\kern-.1667em\lower.7ex\hbox{E}\kern-.125emX}}
\begin{document}

\title{Preserving Near-Optimal Gradient Sparsification Cost for Scalable Distributed Deep Learning}


\author{\IEEEauthorblockN{Daegun Yoon}
\IEEEauthorblockA{\textit{ETRI}\\
\textit{Daejeon, Republic of Korea}\\
kljp@etri.re.kr}
\and
\IEEEauthorblockN{Sangyoon Oh}
\IEEEauthorblockA{\textit{Ajou University}\\
\textit{Suwon, Republic of Korea}\\
syoh@ajou.ac.kr}
}

\maketitle

\begin{abstract}
Communication overhead is a major obstacle to scaling distributed training systems. Gradient sparsification is a potential optimization approach to reduce the communication volume without significant loss of model fidelity. However, existing gradient sparsification methods have low scalability owing to inefficient design of their algorithms, which raises the communication overhead significantly. In particular, gradient build-up and inadequate sparsity control methods degrade the sparsification performance considerably. Moreover, communication traffic increases drastically owing to workload imbalance of gradient selection between workers.

To address these challenges, we propose a novel gradient sparsification scheme called ExDyna. In ExDyna, the gradient tensor of the model comprises fined-grained blocks, and contiguous blocks are grouped into non-overlapping partitions. Each worker selects gradients in its exclusively allocated partition so that gradient build-up never occurs. To balance the workload of gradient selection between workers, ExDyna adjusts the topology of partitions by comparing the workloads of adjacent partitions. In addition, ExDyna supports online threshold scaling, which estimates the accurate threshold of gradient selection on-the-fly. Accordingly, ExDyna can satisfy the user-required sparsity level during a training period regardless of models and datasets. Therefore, ExDyna can enhance the scalability of distributed training systems by preserving near-optimal gradient sparsification cost. In experiments, ExDyna outperformed state-of-the-art sparsifiers in terms of training speed and sparsification performance while achieving high accuracy.
\end{abstract}

\begin{IEEEkeywords}
distributed deep learning, gradient sparsification, scalability
\end{IEEEkeywords}

\section{Introduction}\label{sec:1}
Data-parallel distributed deep learning is the most widely adopted solution to accelerate deep neural network (DNN) model\cite{gpt4,palm2} training. However, the communication overhead becomes the major performance bottleneck as the distributed training system scales out. Gradient sparsification\cite{dls,acpsgd,adacomp,accordion,learnedcomp,natural,errorcompensatedx} is a remedy to reduce the communication overhead by using only the largest $k$ gradients in the entire gradient tensor for model update. In many cases, up to 99.9\% of gradients can be sparsified without significant loss of model fidelity\cite{bbtopk,oktopk}. This sparsity can also be translated to the communication density ($d=\frac{k}{n_g}$), which is the ratio of selected gradients ($k$) to all gradients ($n_g$). Density is an important metric to evaluate the communication efficiency of the sparsifier. Thus, how the top $k$ gradients are selected is critical to the design of a sparsifier, and the simplest way is sorting the entire gradient tensor and extracting the top $k$ elements\cite{gtopk,lagssgd,omgssgd,bbtopk,convproof01,scalecom}.

However, the cost of sorting-based selection is high because the computational complexity of the top-k operation is basically $O({n_g}\log{k})$\cite{topkcomplexity}. This computational cost remains constant even if the distributed training system scales out. To reduce gradient selection cost, a threshold can be used to identify whether each gradient should be selected or not. Threshold-based gradient selection\cite{hardthreshold,sidco,oktopk} incurs comparably lower computational overhead than the sorting-based approach. Moreover, the threshold-based approach can easily be parallelized, which gives the huge advantage of exploiting the massive parallelism of GPU cores.

Despite the low computational cost of gradient selection, threshold-based sparsifiers suffer from the challenge of increased communication cost. In threshold-based sparsifiers, it is challenging to estimate the accurate threshold that satisfies the communication density a user wants to maintain\cite{dgc,sidco,oktopk}. This problem makes the actual density increase significantly. Moreover, most gradient sparsifiers cause the gradient build-up problem\cite{scalecom}, which augments the actual density because the selected gradients of each worker do not fully overlap with those of other workers. Thus, the actual density from the number of aggregated gradients increases up to $n$ times the user-set density when $n$ workers participate in the distributed training.

Furthermore, additional communication overhead occurs owing to the difference in the number of selected gradients between workers\cite{bbtopk}. Gradient sparsification requires the all-gather\footnote{The communication procedure of gradient sparsification consists of a series of all-gather and all-reduce operations whereas non-sparsified training only requires all-reduce operation.} operation to aggregate all selected gradients; however, the communication traffic of all-gather is proportional to the payload of the worker with the largest payload among workers. This is because the payload of each worker must be padded to meet the highest payload size. Thus, communication overhead increases as workload becomes more imbalanced.

\begin{figure*}[t]
    \centering
    \begin{subfigure}[t]{0.321\textwidth}
        \centering
        \includegraphics[width=1.0\linewidth]{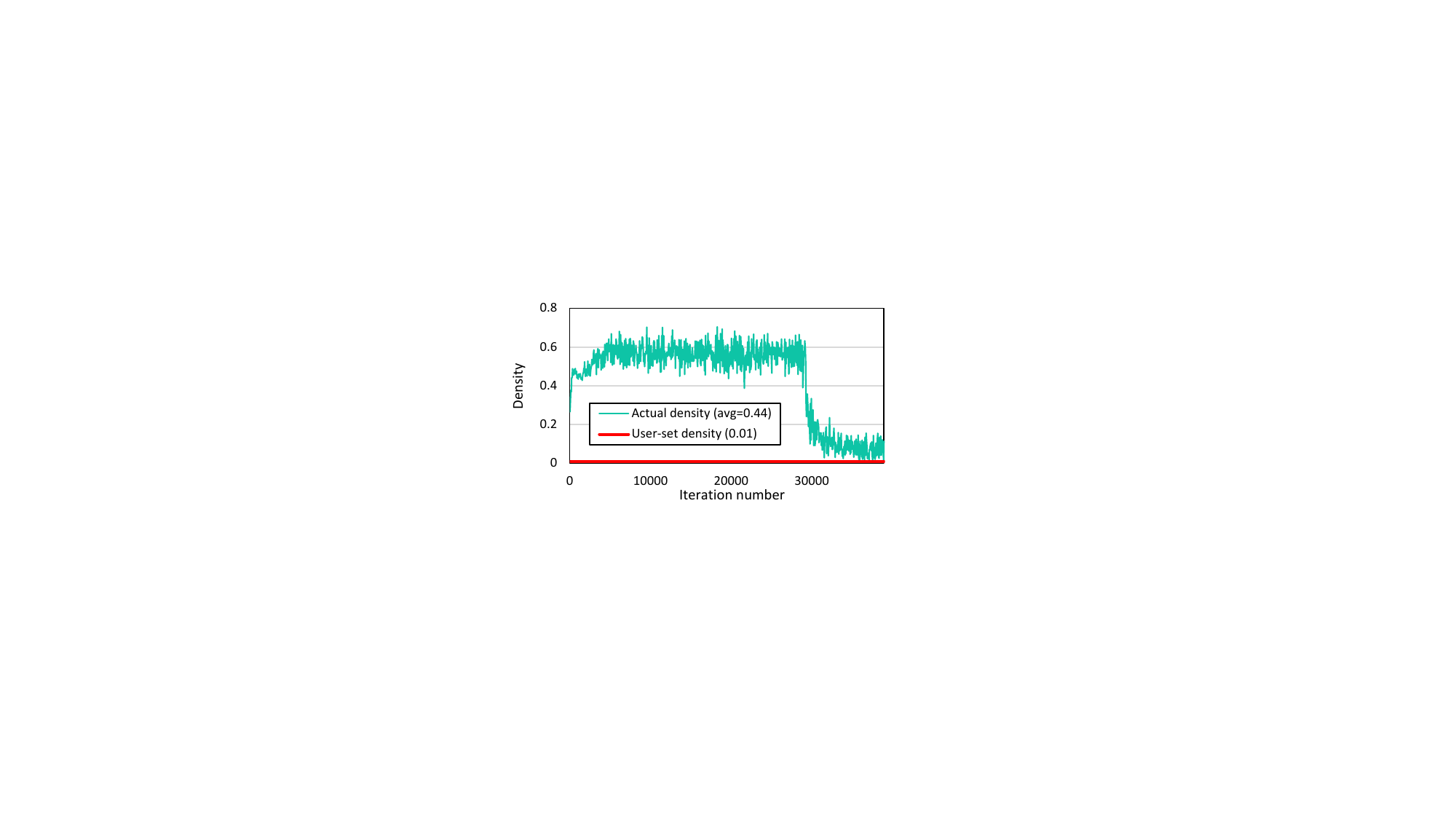}
        \caption{ResNet-18 on CIFAR-100 ($d=0.01$)}
        \label{fig:1a}
    \end{subfigure}
    ~ 
    \begin{subfigure}[t]{0.321\textwidth}
        \centering
        \includegraphics[width=1.0\linewidth]{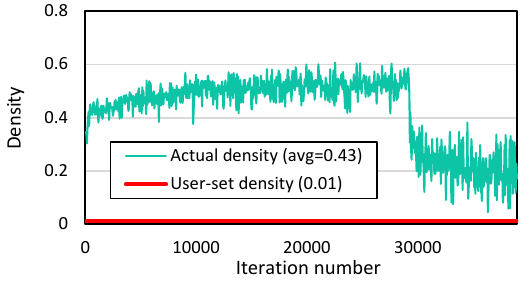}
        \caption{GoogLeNet on CIFAR-100 ($d=0.01$)}
        \label{fig:1b}
    \end{subfigure}
    ~ 
    \begin{subfigure}[t]{0.321\textwidth}
        \centering
        \includegraphics[width=1.0\linewidth]{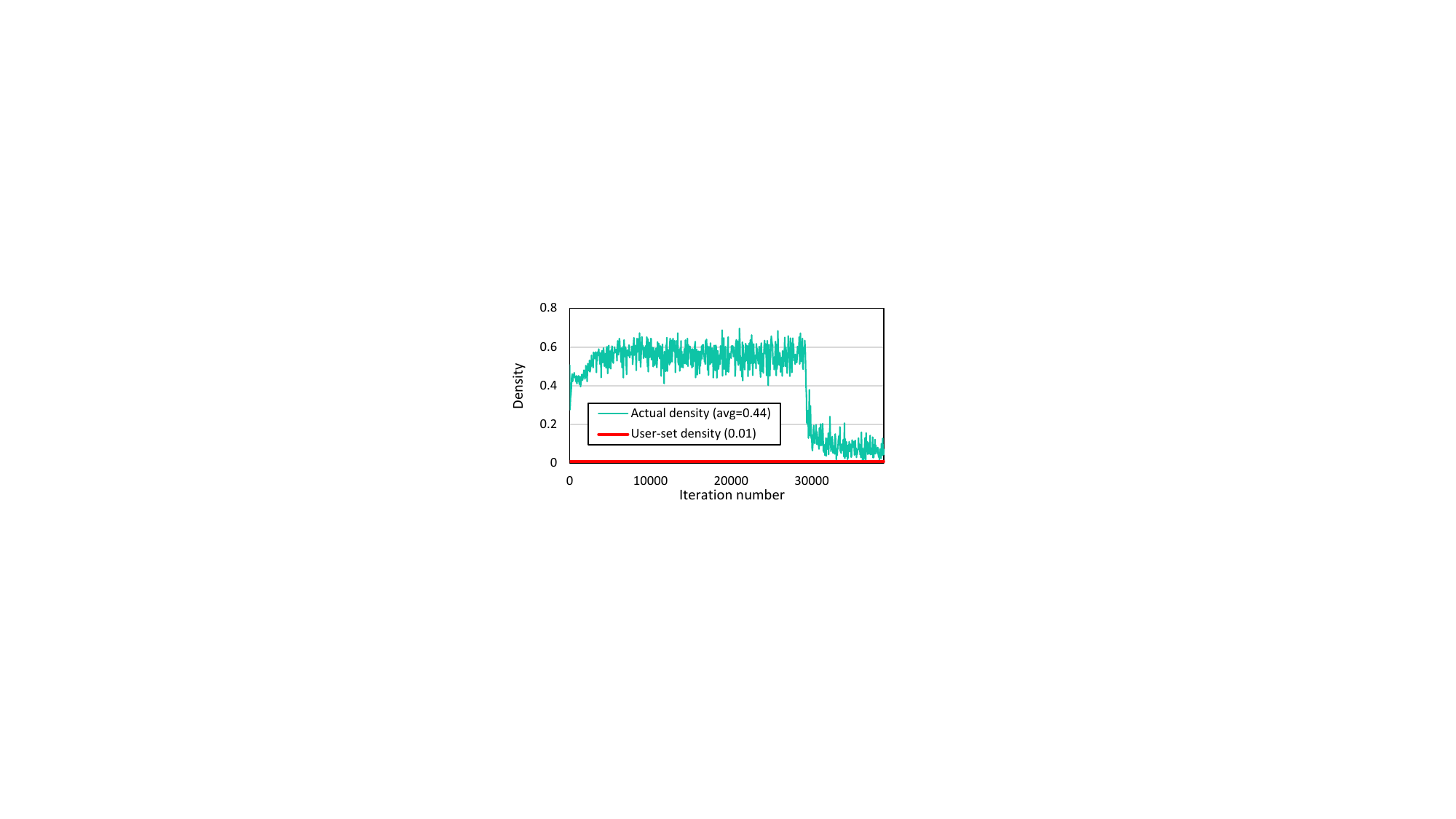}
        \caption{SENet-18 on CIFAR-100 ($d=0.01$)}
        \label{fig:1c}
    \end{subfigure}
    \caption{Challenges in scalable gradient sparsification in terms of communication density increase: gradient build-up and inappropriate threshold estimation. All experiments were conducted on 8 GPUs.}
    \label{fig:1}
\end{figure*}

\begin{figure*}[t]
    \centering
    \begin{subfigure}[t]{0.321\textwidth}
        \centering
        \includegraphics[width=1.0\linewidth]{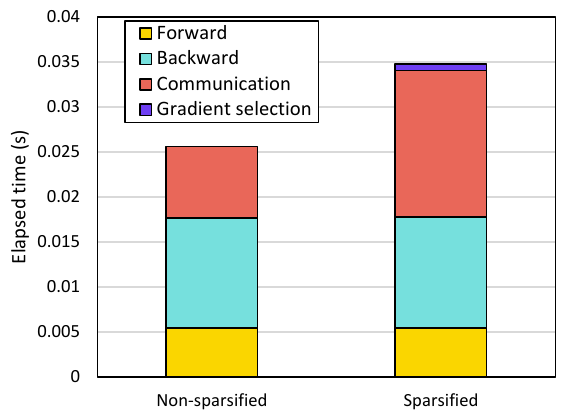}
        \caption{ResNet-18 on CIFAR-100 ($d=0.01$)}
        \label{fig:2a}
    \end{subfigure}
    ~ 
    \begin{subfigure}[t]{0.321\textwidth}
        \centering
        \includegraphics[width=1.0\linewidth]{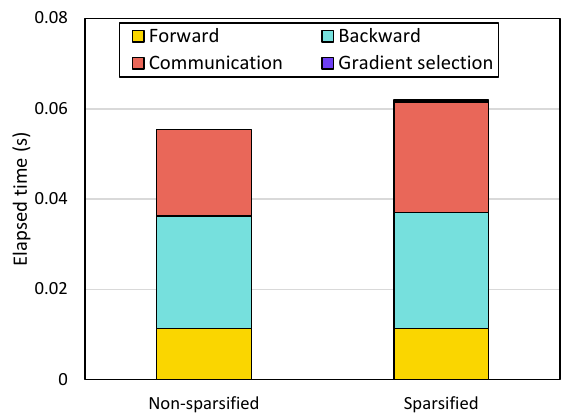}
        \caption{GoogLeNet on CIFAR-100 ($d=0.01$)}
        \label{fig:2b}
    \end{subfigure}
    ~ 
    \begin{subfigure}[t]{0.321\textwidth}
        \centering
        \includegraphics[width=1.0\linewidth]{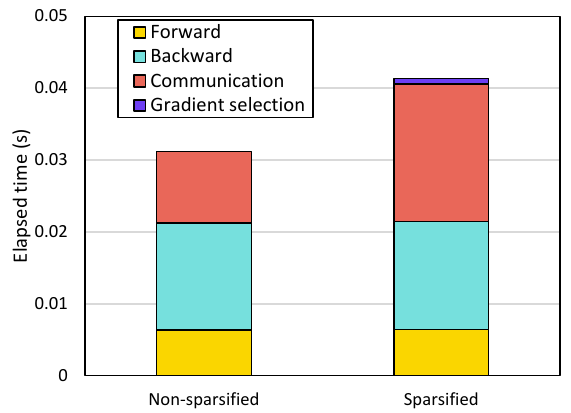}
        \caption{SENet-18 on CIFAR-100 ($d=0.01$)}
        \label{fig:2c}
    \end{subfigure}
    \caption{Communication cost increase of sparsified distributed training owing to challenges: gradient build-up, inaccurate threshold estimation, and workload imbalance. The training time is the average wall-clock time for one iteration. All experiments were conducted on 8 GPUs.}
    \label{fig:2}
\end{figure*}

To identify these challenges, we measured the communication overhead of the threshold-based sparsifier\cite{hardthreshold} on three computer vision applications: ResNet-18\cite{resnet}, GoogLeNet\cite{googlenet}, and SENet-18\cite{senet} on CIFAR-100\cite{cifar}. Figure~\ref{fig:1} shows the challenges in terms of communication density increase, which hinder the scalability of gradient sparsification. Owing to inaccurate threshold estimation and gradient build-up, the actual density excessively increased in every case. As a consequence, the overall performance of sparsified distributed training may be lower than in the non-sparsified case. In Figure~\ref{fig:2}, sparsified training showed longer communication time than non-sparsified training. Because the all-gather operation is used to collect the selected gradients from other workers, the communication overhead of all-gather can be greater than the advantage brought by sparse communication if the number of collected gradients is too large or workload between workers is unbalanced. In every case of Figures~\ref{fig:1} and \ref{fig:2}, the average actual density is excessively high; thus, we cannot expect training time reduction from gradient sparsification.

In this paper, we propose ExDyna\footnote{ExDyna is an acronym for \textbf{ex}ploiting \textbf{dyna}mic gradient sparsification.} to address these challenges for scalability of distributed training systems. To satisfy the communication traffic at the user-set level, ExDyna estimates an accurate threshold for gradient selection and eliminates gradient build-up. To estimate the threshold, ExDyna monitors the actual density and scales the threshold that can minimize the density error\footnote{Density error is defined as the difference between the actual and user-set densities.}. After training begins, ExDyna can accurately find the threshold that satisfies the user-set density within a few iterations. To eliminate gradient build-up, ExDyna divides the gradient vector of the entire model into multiple blocks in a fine-grained manner. Then, multiple contiguous blocks are grouped into one partition, and the entire search space comprises $n$ partitions. Each partition is exclusively allocated to one worker; thus, each worker can have a contiguous search space that does not overlap with those of other workers. Thus, while gradients can be selected within a whole range of the gradient vector, gradient build-up never occurs.

To balance the workload between workers, the search range of each partition is adjusted by comparing its workload (i.e., number of selected gradients within the partition) with those of adjacent partitions at each training iteration. Thus, the communication overhead increase by zero padding in all-gather can be reduced to an acceptable level. By addressing these challenges, ExDyna can preserve near-optimal gradient sparsification cost for scalable data-parallel distributed deep learning.

\begin{table*}
    \centering
    \caption{Strengths and weaknesses of state-of-the-art gradient sparsifiers and the proposed ExDyna.}
    \label{tab:1}
    \tabulinesep = 1mm
    \begin{tabu} to \linewidth {X[1.1,l,m]X[0.5,c,m]X[1.1,c,m]X[0.7,c,m]X[0.7,c,m]X[0.7,c,m]X[0.5,c,m]X[0.8,c,m]X[0.8,c,m]}
        \toprule
        Sparsifier & Gradient build-up & All-gather padding overhead & Inaccurate threshold & Threshold tuning & Model fidelity loss & Worker idling & Gradient selection cost & Additional overhead \\
        \midrule
        Top-k\cite{convproof01} & Yes & No & No & No & No & No & Very high & No \\
        CLT-k\cite{scalecom} & No & No & No & No & Yes & Yes & Very high & No \\
        Hard-threshold\cite{hardthreshold} & Yes & Very high & Yes & Yes & No & No & Very low & No \\
        SIDCo\cite{sidco} & Yes & Very high & No & No & No & No & Very low & Very high \\
        ExDyna & No & Very low & No & No & No & No & Near-zero  & Near-zero \\
        \bottomrule
    \end{tabu}
\end{table*}

This study makes the following contributions:
\begin{itemize}
    \item \textbf{Online threshold scaling}. This prevents the communication traffic from increasing to an unintended level, allowing user-set density to be satisfied. Moreover, this does not depend on the input model and dataset; thus, it can show consistent sparsification performance in various training settings.
    \item \textbf{Block-based gradient vector partitioning}. This eliminates the gradient build-up; thus, the number of selected gradients can be maintained at the user-set level regardless of the number of participating workers. In other words, the scalability of the distributed training system can be improved significantly.
    \item \textbf{Dynamic partition allocation}. This alleviates the workload imbalance between workers; thus, the communication overhead of all-gather can be reduced significantly. This can also improve the scalability of distributed deep learning.
\end{itemize}

The remainder of this paper is organized as follows. Section~\ref{sec:2} presents the preliminaries for this paper. Section~\ref{sec:3} clarifies the limitations of the state-of-the-art gradient sparsifiers. Section~\ref{sec:4} proposes ExDyna, designed to address the challenges stated in this paper. Section~\ref{sec:5} verifies our contributions by empirical comparisons between ExDyna and state-of-the-art gradient sparsifiers. Finally, Section~\ref{sec:6} concludes the paper.

\section{Preliminaries}\label{sec:2}
Gradient sparsification is a lossy communication algorithm in distributed deep learning. Because only a minority of gradients are used to update a model, stochastic gradient descent (SGD) can proceed toward minima in $k$ dimensions among $n_g$ dimensions at each iteration. Thus, some gradients may have no chance to be selected until training terminates because the gradient magnitude varies with the model layer it belongs to\cite{layernorm}. Moreover, naive sparsifiers are vulnerable to converge to local minima. If SGD has already attained local minimum in a particular dimension, it may not have chance to escape from local minimum as long as gradient sparsification is applied because the gradient is too small to be selected. Thus, model fidelity loss is inevitable for naively sparsified models.

To address this problem, recent sparsifiers mostly accumulate the unselected gradients in local memory of each worker\cite{seide2014}. Accordingly, unselected gradients will have a chance to be selected when their accumulated magnitude becomes sufficiently large. That is, a sparsifier can update the model in most dimensions during a training period. Moreover, sparsifiers can be robust to local minima owing to gradient accumulation. This is because SGD has the ability to escape from local minimum when the accumulated value is large enough to be included in the top $k$ gradients. Therefore, model fidelity loss of gradient sparsification can be alleviated.

Let $e_{i,t}$ be an $n_g$-dimensional vector that accumulates unselected gradients, where $i$ and $t$ are worker and iteration numbers, respectively. If each gradient is selected, its corresponding accumulated value is initialized to zero. We refer to the L2-norm of $e_{i,t}$ as local error of each worker, which is denoted by ${\lVert}e_{i,t}{\rVert}$. Therefore, the global error is denoted by
\begin{equation}
    {\lVert}e_{t}{\rVert}=\frac{1}{n}\sum_{i=0}^{n-1}{{\lVert}e_{i,t}{\rVert}}.
\end{equation}
The global error indicates how much less the sparsified model is updated than the non-sparsified model. That is, if the user-set density gets closer to 1.0 (non-sparsified), the global error gets closer to zero. In this study, we utilized global error to evaluate the threshold estimation performance of the proposed ExDyna. We verified how accurately the online threshold scaling of ExDyna traces the variation of global error over training iterations through experiments detailed in Section~\ref{sec:5}.

\section{Limitations of State-of-the-Art Methods}\label{sec:3}
In this section, we discuss the limitations of state-of-the-art gradient sparsifiers, as listed in Table~\ref{tab:1}.

In gradient sparsification, how to select the top $k$ gradients is critical to the computational and communication cost. In terms of computational cost, sparsifiers are categorized into sorting- and threshold-based approaches. Because finding the top $k$ elements in an $n_g$-sized vector basically requires $O({n_g}\log{k})$ computational complexity, the cost increases significantly with the size of the DNN model. The representative sorting-based sparsifier is the Top-k sparsifier\cite{convproof01}. In the Top-k sparsifier, each worker finds the top $k$ gradients among the entire gradient vector, which brings not only high computational cost but also high communication cost owing to gradient build-up.

To eliminate gradient build-up in sorting-based sparsifiers, cyclic local top-k (CLT-k)\cite{scalecom} was proposed. In CLT-k, only one worker (leader) has the authority to select the top $k$ gradients at each iteration. The selected gradients are broadcasted to other workers so that gradient build-up does not occur. However, this delegated top $k$ selection incurs side-effects. Because all workers except the leader must idle until the gradient selection is finished, computational resources are wasted in the distributed training system. Moreover, model fidelity is reduced because only the leader worker has the chance to select the top $k$ gradients from its locally computed gradients at each iteration. That is, each worker must wait $n-1$ iterations per cycle to have authority; thus, its local gradients may become stale.

The computational cost for finding the top $k$ gradients can be reduced considerably using threshold. However, it is challenging to estimate the threshold that satisfies user-required communication density without significant overhead. SIDCo\cite{sidco} estimates the threshold using a statistical model at each iteration. Thus, the threshold changes to satisfy the user-set density as gradient distribution changes over training iterations. However, utilizing gradient distribution incurs too high computational overhead to use in real-world systems. Moreover, it is difficult to use in general-purpose applications because the trainer must prepare the several predefined statistical models that can estimate the accurate threshold.

\begin{algorithm}[t]
\SetAlgoVlined
\SetAlgoCaptionSeparator{}
\SetNlSty{}{}:{}
\PrintSemicolon
\SetKwInput{KwInput}{Input}
\KwInput{$G(\cdot)$: stochastic gradients\newline
Sparsify($\cdot$): threshold-based gradient sparsifier\newline
$n_g$: number of gradients in model\newline
$n_b$: number of blocks in gradient vector
}
\For{worker $i$ ${\in}$ $[0,n$ - $1]$ \textnormal{\textbf{in parallel}}}{
Initialize model $x_0$ ${\in}$ ${\mathbb{R}}^{n_g}$\;
Initialize local error $e_{i,0}$ ${\leftarrow}$ $0^{n_g}$\;
Initialize partial-$k$ vector $k_t$ ${\leftarrow}$ ${\frac{k}{n}}^{n}$\;
Initialize threshold $\delta_{0}$\;
Initialize partitions $parts_{0}$ ${\leftarrow}$ Partition($n_g$, $n_b$)\;
\For{iteration $t$ ${\geq}$ $0$}{
${acc}_{i,t}$ ${\leftarrow}$ $e_{i,t}$ + ${\eta}_{t}G_{i,t}(x_t)$\;
${part}_{i,t}$, $parts_{t+1}$ ${\leftarrow}$ Allocate($k_t$, $parts_{t}$)\;
${idx}_{i,t}$ ${\leftarrow}$ Sparsify(${acc}_{i,t}$, ${part}_{i,t}$, $\delta_{t}$)\;
${idx}_t$, $k_t$ ${\leftarrow}$ AllGather(${idx}_{i,t}$)\;
$g_{i,t}$ ${\leftarrow}$ ${acc}_{i,t}[{idx}_t]$\;
$g_t$ ${\leftarrow}$ AllReduce($g_{i,t}$, SUM)\;
$k^{\prime}$ ${\leftarrow}$ ${k_t}$.sum()\;
$\delta_{t+1}$ ${\leftarrow}$ Estimate($k$, $k^{\prime}$, $\delta_{t}$)\;
$k_{t+1}$ ${\leftarrow}$ $k_t$\;
$x_{t+1}$ ${\leftarrow}$ $x_t$ - $\frac{1}{n}g_t$\;
${acc}_{i,t}[{idx}_t]$ ${\leftarrow}$ $0$\;
$e_{i,t+1}$ ${\leftarrow}$ ${acc}_{i,t}$\;
}
}
\caption{Distributed SGD with ExDyna}
\label{alg:1}
\end{algorithm}

Meanwhile, the hard-threshold sparsifier\cite{hardthreshold} uses a fixed threshold that is determined before training begins. Accordingly, this sparsifier does not consider the real-time workload and cannot satisfy the user-required communication traffic. Therefore, the hard-threshold sparsifier requires a number of rigorous threshold tuning tasks to maintain the appropriate density for each model and dataset. Moreover, threshold-based sparsifiers\cite{hardthreshold,sidco} entail gradient build-up and all-gather padding overhead, as described in Section~\ref{sec:1}. These problems significantly decrease communication efficiency; thus, the scalability of the distributed training system is also limited.

The limitations of state-of-the-art gradient sparsifiers show that it is very challenging to satisfy all the criteria listed in Table~\ref{tab:1}. In this study, we addressed these challenges using a novel method that preserves near-optimal gradient sparsification cost for scalable distributed deep learning.

\begin{algorithm}[t]
\SetAlgoVlined
\SetAlgoCaptionSeparator{}
\SetNlSty{}{}:{}
\PrintSemicolon
\SetKwInput{KwInput}{Input}
\SetKwInput{KwOutput}{Output}
\KwInput{$n_g$: number of gradients in model\newline
$n_b$: number of blocks in gradient vector
}
\KwOutput{
parts: partitions of gradient vector
}
temp ${\leftarrow}$ $n_g$ / $n_b$\;
sz\_blk ${\leftarrow}$ temp - temp \% 32\;
quotient ${\leftarrow}$ $n_b$ / $n$\;
remainder ${\leftarrow}$ $n_b$ \% $n$\;
blk\_part ${\leftarrow}$ empty($n$)\;
blk\_pos ${\leftarrow}$ zeros($n$)\;
\For{$i$ in range($n$)}{
\If{$i$ ${\in}$ range(remainder)}{
blk\_part[i] ${\leftarrow}$ quotient + $1$\;
}
\Else{
blk\_part[i] ${\leftarrow}$ quotient\;
}
}
\For{$i$ in range($1$, $n$)}{
blk\_pos[i] ${\leftarrow}$ blk\_part[i - 1] + blk\_pos[i - 1]\;
}
parts ${\leftarrow}$ pack(blk\_part, blk\_pos)\;
\caption{Block-based gradient vector partitioning}
\label{alg:2}
\end{algorithm}

\section{ExDyna Design}\label{sec:4}
We designed ExDyna as a gradient sparsifier that divides the gradient vector into fine-grained blocks and exclusively allocates them to workers to enable parallelized gradient selection. ExDyna comprises the following sequence: 1) block-based gradient vector partitioning, 2) dynamic partition allocation, 3) partition-wise exclusive gradient selection, and 4) online threshold scaling. Algorithm~\ref{alg:1} presents the pseudocode for the ExDyna workflow in a distributed training system. These processes in ExDyna are executed after backward propagation at each iteration. The following subsections present a detailed discussion of each process in ExDyna.

\subsection{Block-based gradient vector partitioning}\label{sec:4.1}
To prevent gradient build-up, a gradient sparsifier should guarantee that workers do not select gradients from a search space overlapped with other workers. ExDyna provides workers with a non-overlapping search space via block-based gradient vector partitioning, which equally divides the entire vector into $n_b$ $b$-sized blocks\footnote{For simplicity, we assume that this partitioning does not leave a remainder. However, we do consider the remainder in our implementation.}. Thus, ${n_g}={n_b}{\cdot}{b}$ is established. ExDyna groups the contiguous blocks into a partition. This contiguous grouping enables coalesced memory access\cite{profcuda} in gradient selection. Accordingly, the entire gradient vector comprises $n$ partitions that are not overlapped with each other.

Because the partitions are based on blocks, the topology of partitions can be reconfigured flexibly by adjusting the number of blocks in each partition. This block-based partitioning is advantageous for workload balance over coarse-grained partitioning, which uses $n$ equally divided partitions at every iteration. This is because the coarse-grained partitioning should use the static topology even if the workload of gradient selection is imbalanced between workers.

Algorithm~\ref{alg:2} presents the pseudocode for the block-based gradient vector partitioning. From lines 7 to 14, each partition determines the index of its first block and the number of blocks that belong to the partition. That is, each partition indicates the search range in the entire gradient vector. Algorithm~\ref{alg:2} is only executed before the first iteration begins because the purpose of this routine is to initialize the topology of partitions.

\begin{figure}
\centering
 \includegraphics[width=1.0\linewidth]{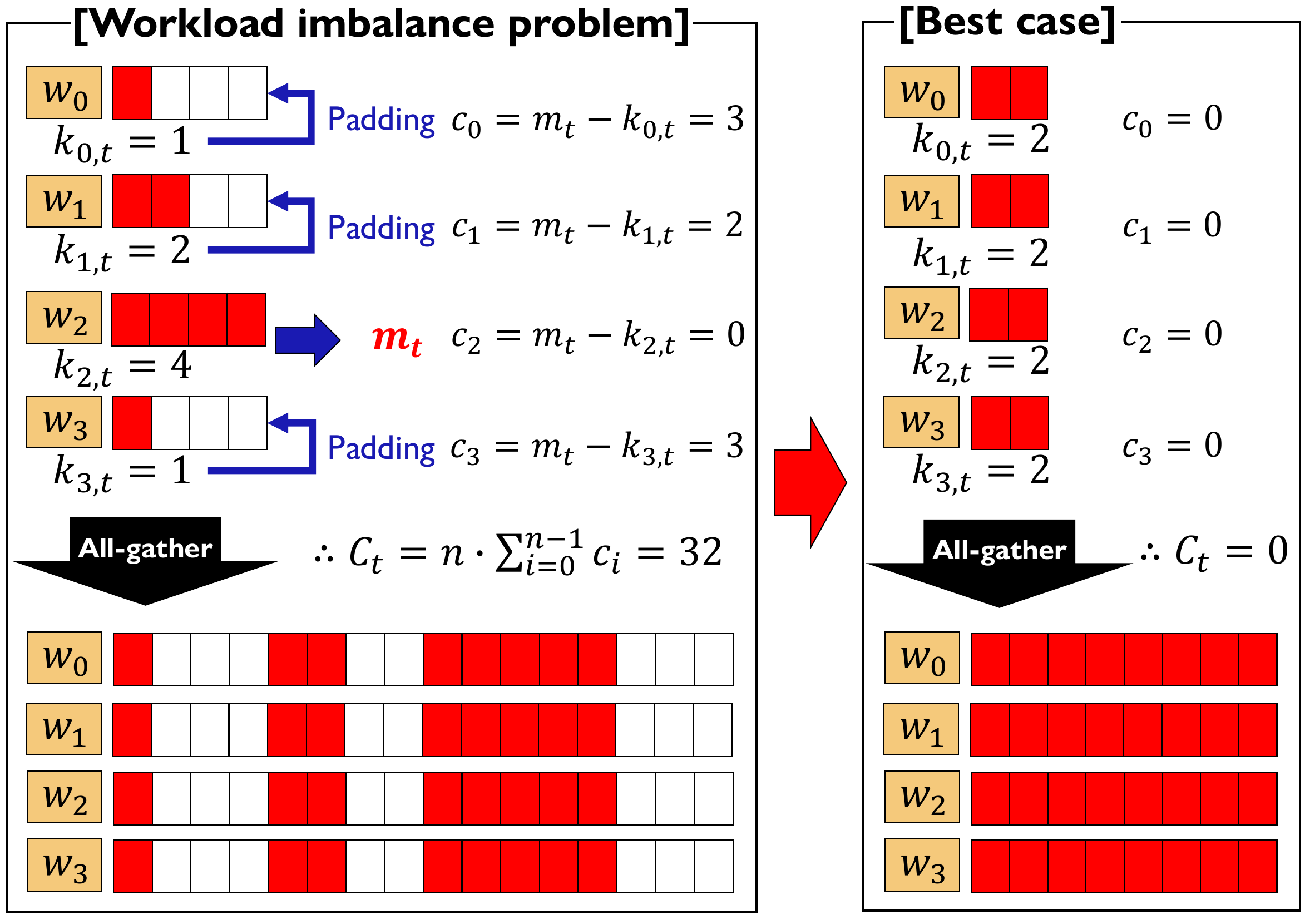}
\caption{Communication overhead increase problem in all-gather operation caused by workload imbalance of gradient selection between workers. The best case indicates that workload is perfectly balanced between workers.}
\label{fig:3}
\end{figure}

\begin{figure}
\centering
 \includegraphics[width=1.0\linewidth]{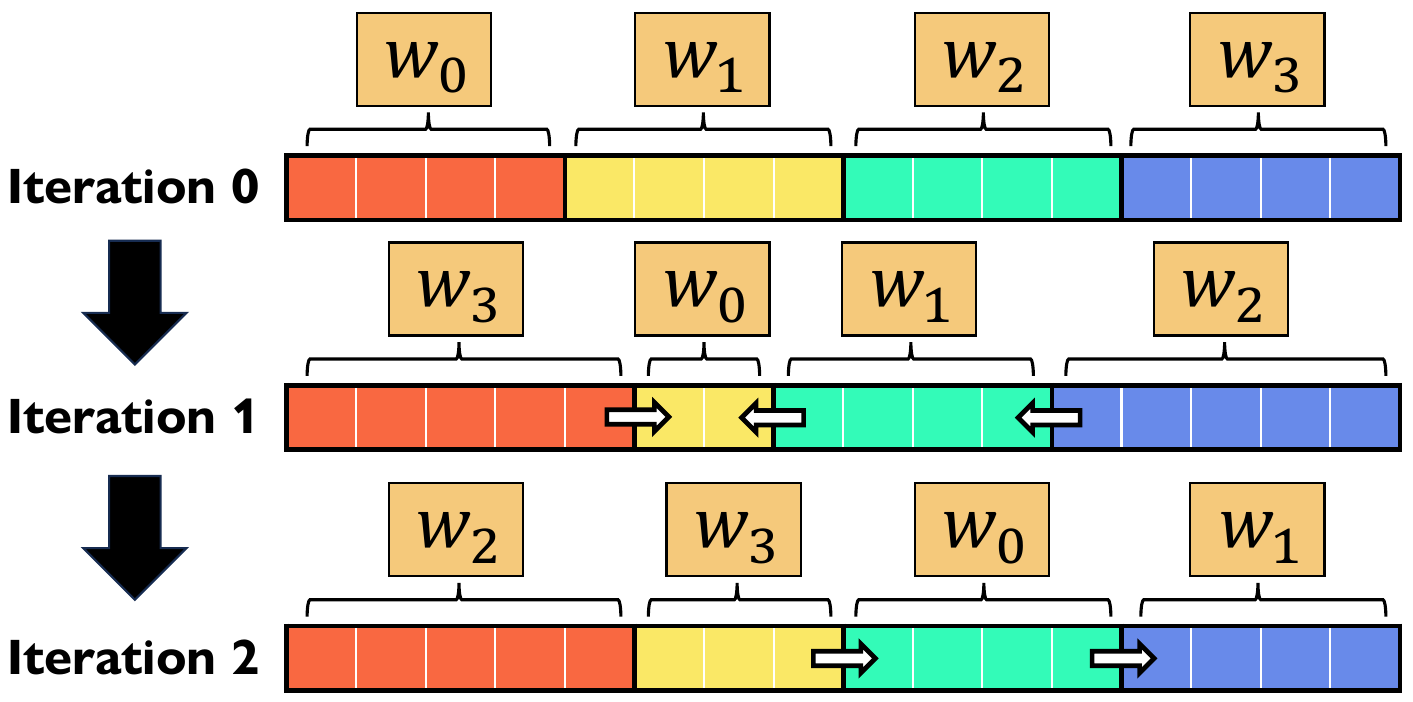}
\caption{Example of dynamic partition allocation. Each square in a partition is a block that contains the fixed number of gradients.}
\label{fig:4}
\end{figure}

\begin{algorithm}[t]
\SetAlgoVlined
\SetAlgoCaptionSeparator{}
\SetNlSty{}{}:{}
\PrintSemicolon
\SetKwInput{KwInput}{Input}
\SetKwInput{KwOutput}{Output}
\KwInput{$k_t$: partial-$k$ vector\newline
parts: partitions of gradient vector
}
\KwOutput{part: allocated partition\newline
parts: adjusted partitions of gradient vector
}
blk\_part, blk\_pos ${\leftarrow}$ parts.unpack()\;
temp ${\leftarrow}$ zeros($n$)\;
\For{$i$ in range($n$)}{
j ${\leftarrow}$ (($t$ - 1) \% $n$ + $i$) \% $n$\;
temp[j] ${\leftarrow}$ $k_t$[i]\;
}
$k_t$ ${\leftarrow}$ temp\;
pk\_prev ${\leftarrow}$ $k_t$.sum() / $n$\;
den\_prev ${\leftarrow}$ $k_t$.sum() / $n_g$\;
\For{$i$ in range($n$ - $1$)}{
det ${\leftarrow}$ $k_t$[i] / pk\_prev\;
det2 ${\leftarrow}$ $k_t$[i + 1] / pk\_prev\;
k\_move ${\leftarrow}$ blk\_move $\cdot$ sz\_blk $\cdot$ den\_prev\;
\If{det $>$ $\alpha$ and det2 $<$ $\frac{1}{\alpha}$}{
\If{blk\_part[i] - blk\_move $<$ min\_blk}{
continue\;
}
blk\_part[i] -= blk\_move\;
blk\_part[i + 1] += blk\_move\;
blk\_pos[i + 1] -= blk\_move\;
$k_t$[i] -= k\_move\;
$k_t$[i + 1] += k\_move\;
}
\If{det $<$ $\frac{1}{\alpha}$ and det2 $>$ $\alpha$}{
\If{blk\_part[i + 1] - blk\_move $<$ min\_blk}{
continue\;
}
blk\_part[i] += blk\_move\;
blk\_part[i + 1] -= blk\_move\;
blk\_pos[i + 1] += blk\_move\;
$k_t$[i] += k\_move\;
$k_t$[i + 1] -= k\_move\;
}
}
alloc ${\leftarrow}$ ($t$ \% $n$ + rank) \% $n$\;
st\_part ${\leftarrow}$ blk\_pos[alloc] $\cdot$ sz\_blk\;
end\_part ${\leftarrow}$ (blk\_pos[alloc] + blk\_part[alloc]) $\cdot$ sz\_blk\;
part ${\leftarrow}$ pack(st\_part, end\_part)\;
parts ${\leftarrow}$ pack(blk\_part, blk\_pos)\;
\caption{Dynamic partition allocation}
\label{alg:3}
\end{algorithm}

\subsection{Dynamic partition allocation}\label{sec:4.2}
To alleviate the workload imbalance of gradient selection between workers, a threshold-based sparsifier should reduce the number of padded elements in the all-gather operation. Figure~\ref{fig:3} shows the communication overhead increase problem in the all-gather operation caused by workload imbalance of gradient selection between workers. Let $k_{i,t}$ be the number of gradients selected in worker $i$'s partition at iteration $t$. Then, the largest number of selected gradients among $n$ workers is denoted by
\begin{equation}
    {m_t}={\max_{i{\in}[0,n-1]}{k_{i,t}}}.
\end{equation}
For all-gather, each worker must pad zero elements to match the worker with the largest $k_{i,t}$. Thus, the number of padded elements for worker $i$ is denoted by
\begin{equation}
    {c_i}={m_t}-{k_{i,t}}={\max_{i{\in}[0,n-1]}{k_{i,t}}}-{k_{i,t}}.
\end{equation}
Accordingly, the communication overhead by workload imbalance between workers in iteration $t$ is formulated as follows:
\begin{equation}
    {C_t}={n}{\cdot}\sum_{i=0}^{n-1}{c_i}={n}{\cdot}\sum_{i=0}^{n-1}{\Big(\max_{i{\in}[0,n-1]}{k_{i,t}}}-{k_{i,t}\Big)}.
\end{equation}

\begin{algorithm}[t]
\SetAlgoVlined
\SetAlgoCaptionSeparator{}
\SetNlSty{}{}:{}
\PrintSemicolon
\SetKwInput{KwInput}{Input}
\SetKwInput{KwOutput}{Output}
\KwInput{$g$: gradient vector\newline
part: index range of allocated partition\newline
$\delta_{t}$: previous threshold
}
\KwOutput{
part\_idx: partially selected gradient indices
}
st\_part, end\_part ${\leftarrow}$ part.unpack()\;
part\_idx ${\leftarrow}$ where($g$[st\_part:end\_part].abs() ${\geq}$ ${\delta_{t}}$)\;
part\_idx ${\leftarrow}$ part\_idx + st\_part\;
\caption{Partition-wise gradient selection}
\label{alg:4}
\end{algorithm}

The communication overhead $C_t$ can be reduced to zero if the workload of gradient selection is perfectly balanced between workers. We refer to this case as the best case in the all-gather operation for gradient aggregation. Then, we quantify how much the communication traffic has increased compared to the best case. We express this increase as a percentage as follows:
\begin{equation}
    f(t)=\frac{{n}{\cdot}{m_t}}{k^{\prime}}=\frac{{n}{\cdot}{\max_{i{\in}[0,n-1]}{k_{i,t}}}}{\sum_{i=0}^{n-1}{k_{i,t}}}.
\end{equation}

The dynamic partition allocation of ExDyna is focused on minimizing $f(t)$ by adjusting $k_{i,t}$ of each partition. Figure~\ref{fig:4} shows an example of this process. Prior to allocating partitions to workers, the topology of partitions should be updated. For this, each two adjacent partitions are inspected to determine whether their gradients are more/less selected than ${k^{\prime}}/{n}$. If one partition selected more gradients and another selected less gradients than ${k^{\prime}}/{n}$, a fixed number of blocks are moved from the former to the latter partition. If this condition is not met, the adjacent partitions are not adjusted. This process is performed iteratively until every two adjacent partitions are inspected. Then, each partition is exclusively allocated to one worker in cyclic order. For example, if the partition $0$ is allocated to worker $0$ in the current iteration, the partition $1$ will be allocated to that worker in the next iteration.

Algorithm~\ref{alg:3} presents the pseudocode for the dynamic partition allocation. From lines 13 to 20, ExDyna determines whether blocks should be moved from the left to right partition. From lines 21 to 28, ExDyna also determines whether blocks should be moved from the right to left partition. Therefore, partitions are not adjusted if the workload is balanced between the partitions. From lines 29 to 31, the partitions are allocated to the workers in cyclic order. The computational overhead of dynamic partition allocation is negligible because the main part of the algorithm (from lines 9 to 28) is based on modifying the several elements of $n$-sized arrays. That is, the complexity of the algorithm only depends on the number of workers instead of the size of the model. Therefore, the overhead remains constant when the number of workers is fixed, even if the size of the model increases.

\begin{algorithm}[t]
\SetAlgoVlined
\SetAlgoCaptionSeparator{}
\SetNlSty{}{}:{}
\PrintSemicolon
\SetKwInput{KwInput}{Input}
\SetKwInput{KwOutput}{Output}
\KwInput{$k$: user-set number of selected gradients\newline
$k^{\prime}$: actual number of selected gradients\newline
$\delta_{t}$: previous threshold
}
\KwOutput{
$\delta_{t+1}$: next threshold
}
exam ${\leftarrow}$ $\frac{k^{\prime}}{k}$\;
\If{exam $>$ $\beta$}{
sf ${\leftarrow}$ $1$ + $\gamma$\;
}
\ElseIf{exam $>$ $\frac{1}{\beta}$}{
sf ${\leftarrow}$ $1$ + $\frac{1}{4}$ {$\cdot$} $\gamma$\;
}
\Else{
sf ${\leftarrow}$ $1$ - $\gamma$\;
}
$\delta_{t+1}$ ${\leftarrow}$ $\delta_{t}$ {$\cdot$} sf\;
\caption{Online threshold scaling}
\label{alg:5}
\end{algorithm}

\subsection{Partition-wise exclusive gradient selection}\label{sec:4.3}
Based on the topology of adjusted partitions, ExDyna performs partition-wise gradient selection. Each worker exclusively selects the gradients in its allocated partition. Thus, the gradient selection task is parallelized across the workers. This parallelization does not cause model fidelity loss because the entire gradient vector is inspected by all workers.

Because each partition is a contiguous range of the entire gradient vector, the partition-wise gradient selection has advantages in the characteristics of the GPU architecture. In terms of global memory access pattern, the contiguous vector range enables coalesced memory access\cite{profcuda}; thus, the number of accesses to the GPU global memory can be minimized. Moreover, memory divergence\cite{memorydiv} does not occur. In addition, gradient selection can be accelerated further by exploiting the parallelism of warp-based SIMD execution. Therefore, the partition-wise exclusive gradient selection can be performed with near-zero computational cost.

Algorithm~\ref{alg:4} presents the pseudocode for the partition-wise exclusive gradient selection. In line 2, all contiguous gradients within the partition are examined by the threshold.

\subsection{Online threshold scaling}\label{sec:4.4}
To satisfy the user-set density, the threshold should be estimated accurately. Instead of deriving a new threshold using a computation-intensive method for each iteration, ExDyna focuses on minimizing the density error by scaling the threshold on-the-fly. The density error in iteration $t$ is denoted by
\begin{equation}
    {\epsilon}_t=|d-d^{\prime}|=\frac{|k-k^{\prime}|}{n_g}=\frac{|k-\sum_{i=0}^{n-1}{k_{i,t}}|}{n_g}.
\end{equation}

To make ${\epsilon}_t$ close to zero, ExDyna scales the next threshold by multiplying the scaling factor with the previous threshold. In each iteration, the actual and user-set densities are compared to decide whether the threshold should increase or decrease. If more gradients were selected than the user required, the threshold should increase to reduce the actual density of the next iteration. On the contrary, the threshold should decrease to raise the actual density if a sparsifier selected a smaller number of gradients than the user required. As the iterations proceed, the threshold converges to a certain value; thus, ${\epsilon}_t$ also approaches zero.

\begin{figure*}[t]
    \centering
    \begin{subfigure}[t]{0.314\textwidth}
        \centering
        \includegraphics[width=1.0\linewidth]{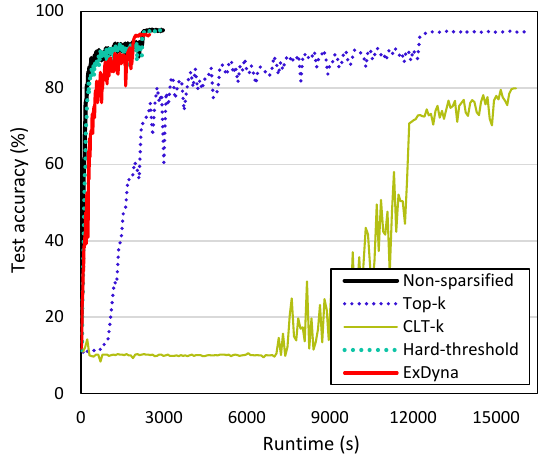}
        \caption{ResNet-152 on CIFAR-10 ($d=0.001$)}
        \label{fig:5a}
    \end{subfigure}
    ~ 
    \begin{subfigure}[t]{0.326\textwidth}
        \centering
        \includegraphics[width=1.0\linewidth]{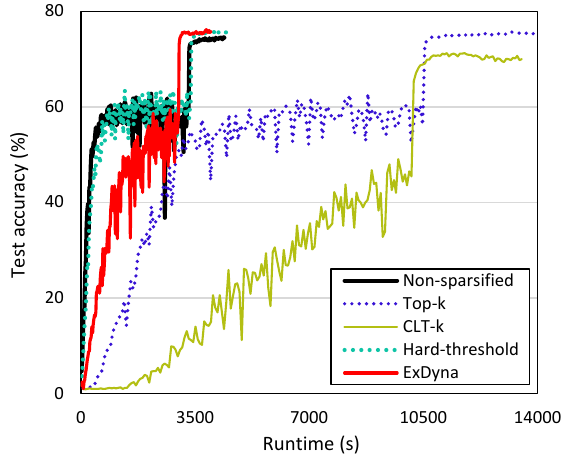}
        \caption{Inception-v4 on CIFAR-100 ($d=0.001$)}
        \label{fig:5b}
    \end{subfigure}
    ~ 
    \begin{subfigure}[t]{0.324\textwidth}
        \centering
        \includegraphics[width=1.0\linewidth]{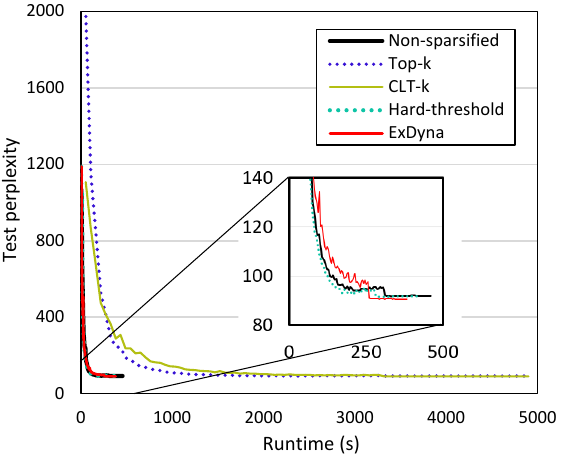}
        \caption{LSTM on WikiText-2 ($d=0.001$)}
        \label{fig:5c}
    \end{subfigure}
    \caption{Convergence performance of sparsified- and non-sparsified- distributed training on 16 GPUs.}
    \label{fig:5}
\end{figure*}

\begin{table}
    \centering
    \caption{Description of each DNN application. $B_l$: local batch size, $n_e$: number of epochs.}
    \label{tab:2}
    \begin{tabular}{lllll}
        \toprule
        Application & Model & Dataset & $B_l$ & $n_e$ \\
        \midrule
        Computer vision & ResNet-152 & CIFAR-10 & 32 & 200\\
        Computer vision & Inception-v4 & CIFAR-100 & 32 & 200\\
        Language modeling & LSTM & WikiText-2 & 32 & 90\\
        \bottomrule
    \end{tabular}
\end{table}

This convergence of ${\epsilon}_t$ is achievable owing to the fine-tuning of the threshold. This is because ${\epsilon}_t$ cannot approach zero if the threshold changes coarsely. Moreover, because the global error ${\lVert}e_{t}{\rVert}$ is reduced as the model comes close to convergence, it is adequate to scale the threshold finely to maintain the accurately estimated threshold. Therefore, ExDyna can satisfy the user-set density during a training period owing to the online threshold scaling.

Algorithm~\ref{alg:5} presents the pseudocode for the online threshold scaling. From lines 2 to 7, the scaling factor of the threshold is determined by inspecting whether the gradients are more or less selected than user required. In line 8, the next threshold is derived by fine-tuning based on the scaling factor.

\section{Evaluation}\label{sec:5}
\subsection{Methodology}\label{sec:5.1}
\textbf{System configuration}. We conducted all experiments on a cluster comprising two GPU nodes. Each node is equipped with eight NVIDIA Tesla V100 GPUs with NVLink, two 16-core Intel Xeon Gold 6226R @ 2.90 GHz CPUs, and 384 GB DDR4 memory. That is, experiments were conducted with 16 GPUs; each case of scalability evaluation used 2, 4, 8, and 16 GPUs. For distributed training, mpirun of OpenMPI 4.0.5\cite{openmpidoc} was used for multiprocess execution of workers, where each worker was run on one GPU with CUDA 10.1\cite{cudadoc}.

\textbf{Models and datasets}. We evaluated the performance of ExDyna and other sparsifiers (Top-k, CLT-k, and hard-threshold sparsifiers) on three DNN applications: 1) ResNet-152\cite{resnet} on CIFAR-10\cite{cifar}, 2) Inception-v4\cite{inceptionv4} on CIFAR-100, and long short-term memory (LSTM)\cite{lstm} on WikiText-2\cite{wikitext}. Table~\ref{tab:2} describes each DNN application used in our evaluation.

\textbf{Implementation}. ExDyna and other sparsifiers were implemented using the deep learning framework PyTorch 1.5\cite{pytorch}, and NCCL 2.4\cite{nccl} was adopted as the communication backend for multi-node multi-GPU distributed training by supporting broadcast, all-gather, and all-reduce primitives. The source code includes everything required to reproduce the experimental results of this paper and is publicly available at \url{https://github.com/kljp/exdyna}.

\textbf{Metrics}. The metrics used for performance evaluation are as follows:
\begin{itemize}
    \item Convergence performance: The test accuracy or perplexity by wall-clock time was measured to evaluate how fast each sparsifier attained convergence.
    \item Sparsification performance: The actual density was measured to evaluate how well each sparsifier satisfied the user-set density.
    \item End-to-end training performance: The training time for one iteration was measured in wall-clock time and is shown as a breakdown to evaluate the computational and communication overheads.
\end{itemize}

\subsection{Performance evaluation}\label{sec:5.2}
\textbf{Convergence performance}. Figure~\ref{fig:5} shows the convergence performance comparison between ExDyna and other approaches, including non-sparsified distributed training. In every application, ExDyna finishes the training in the shortest time among all approaches while attaining a convergence point close to that of non-sparsified training. The hard-threshold sparsifier also shows significant performance, while its convergence rate is higher than that of ExDyna. This is because the hard-threshold sparsifier selects a significantly larger number of gradients than ExDyna. Accordingly, the hard-threshold sparsifier shows slower training speed than ExDyna.

Meanwhile, sorting-based sparsifiers (Top-k and CLT-k) were incomparably slower than threshold-based sparsifiers owing to their high computational cost of gradient selection based on the top-k operation. Moreover, CLT-k shows a significantly low convergence rate owing to the loss of model fidelity caused by delegated gradient selection.

\begin{figure*}[t]
    \centering
    \begin{subfigure}[t]{0.314\textwidth}
        \centering
        \includegraphics[width=1.0\linewidth]{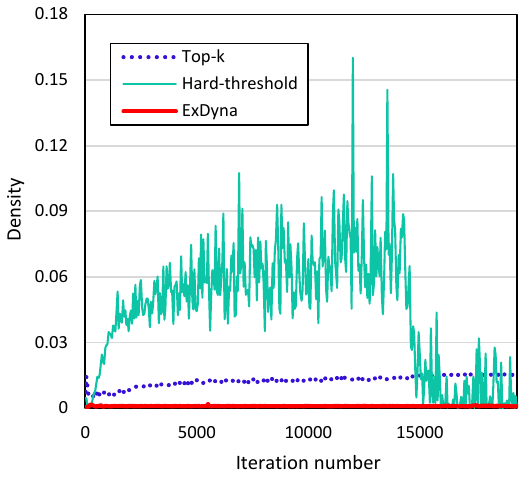}
        \caption{ResNet-152 on CIFAR-10 ($d=0.001$)}
        \label{fig:6a}
    \end{subfigure}
    ~ 
    \begin{subfigure}[t]{0.314\textwidth}
        \centering
        \includegraphics[width=1.0\linewidth]{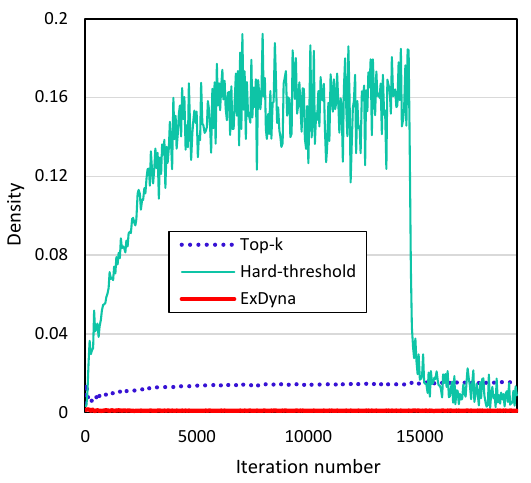}
        \caption{Inception-v4 on CIFAR-100 ($d=0.001$)}
        \label{fig:6b}
    \end{subfigure}
    ~ 
    \begin{subfigure}[t]{0.329\textwidth}
        \centering
        \includegraphics[width=1.0\linewidth]{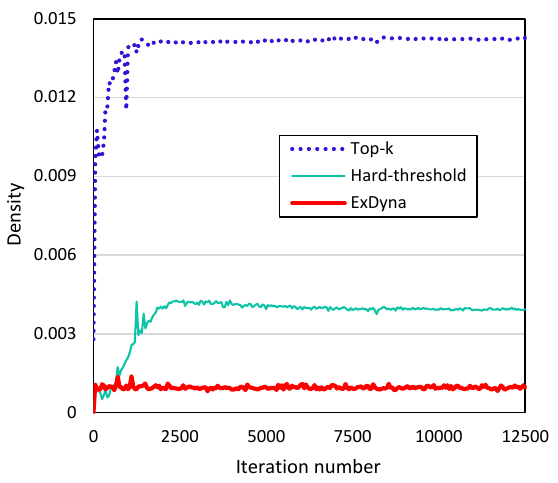}
        \caption{LSTM on WikiText-2 ($d=0.001$)}
        \label{fig:6c}
    \end{subfigure}
    \caption{Sparsification performance of sparsifiers on 16 GPUs. The Y-axis indicates the actual density measured over training iterations.}
    \label{fig:6}
\end{figure*}

\begin{figure*}[t]
    \centering
    \begin{subfigure}[t]{0.319\textwidth}
        \centering
        \includegraphics[width=1.0\linewidth]{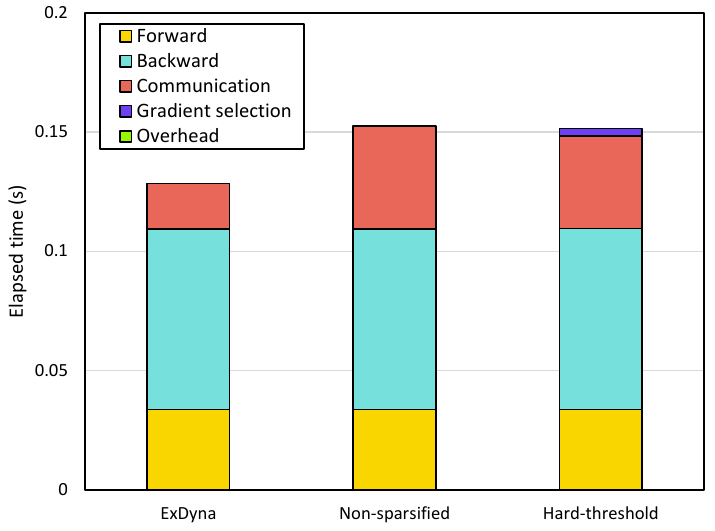}
        \caption{ResNet-152 on CIFAR-10 ($d=0.001$)}
        \label{fig:7a}
    \end{subfigure}
    ~ 
    \begin{subfigure}[t]{0.320\textwidth}
        \centering
        \includegraphics[width=1.0\linewidth]{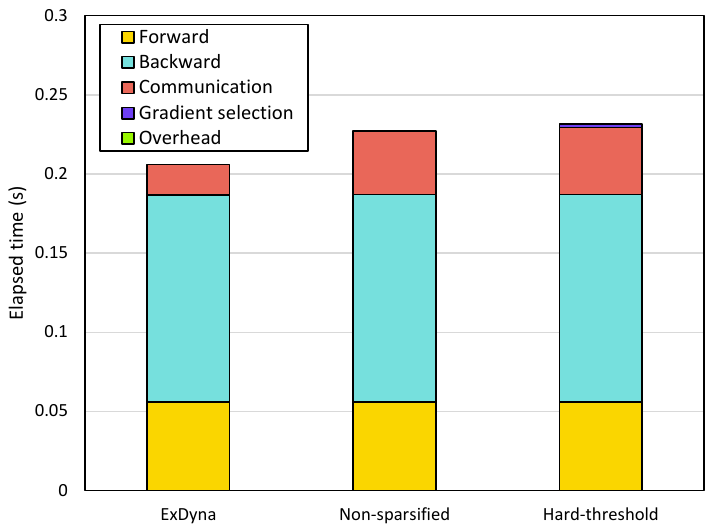}
        \caption{Inception-v4 on CIFAR-100 ($d=0.001$)}
        \label{fig:7b}
    \end{subfigure}
    ~ 
    \begin{subfigure}[t]{0.322\textwidth}
        \centering
        \includegraphics[width=1.0\linewidth]{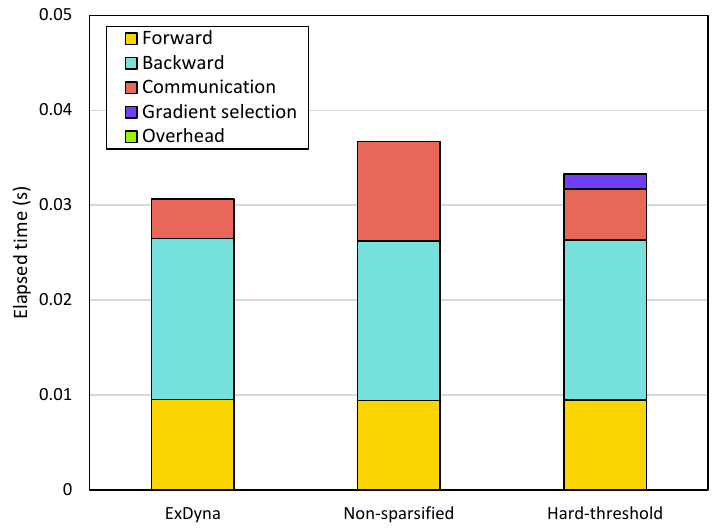}
        \caption{LSTM on WikiText-2 ($d=0.001$)}
        \label{fig:7c}
    \end{subfigure}
    \caption{Training time breakdown of threshold-based sparsifiers and non-sparsified distributed training on 16 GPUs. The training time is the average wall-clock time for one iteration.}
    \label{fig:7}
\end{figure*}

\textbf{Sparsification performance}. Figure~\ref{fig:6} shows the sparsification performance comparison between ExDyna, hard-threshold, and Top-k sparsifiers. In every experiment, ExDyna maintained the actual density at the user-set density of 0.001. On the contrary, the hard-threshold shows significantly increased actual density owing to gradient build-up and an inaccurate threshold in every application. In particular, the average actual density of hard-threshold was 106.6$\times$ the user-set density on Inception-v4. This density increase is the reason why the hard-threshold sparsifier shows higher convergence rate and slower training time than ExDyna. In computer vision applications, the density of hard-threshold sparsifier suddenly drops after iteration 14,600, where the learning rate decay is activated. This drop is because the global error is not large owing to the almost converged model.

\textbf{End-to-end training performance}. Figure~\ref{fig:7} shows the breakdown of training time for one iteration. In every experiment, ExDyna showed the fastest training time among all approaches. This is because ExDyna significantly reduces computational and communication costs, which are major performance bottlenecks in existing sparsifiers. Meanwhile, the hard-threshold sparsifier can only alleviate the computational overhead. Owing to gradient build-up and an inaccurate threshold, the hard-threshold sparsifier cannot resolve the communication overhead. This overhead will increase significantly if the user-set density is higher.

For CLT-k, the training times were 6.31$\times$, 3.38$\times$, and 12.79$\times$ higher than those of ExDyna on ResNet-152, Inception-v4, and LSTM, respectively. Similarly, the training times of Top-k were 6.51$\times$, 3.50$\times$, and 12.85$\times$ higher than those of ExDyna on the three DNN applications, respectively. Therefore, the use of sorting-based sparsifiers is impractical.

\begin{figure*}[t]
    \centering
    \begin{subfigure}[t]{0.319\textwidth}
        \centering
        \includegraphics[width=1.0\linewidth]{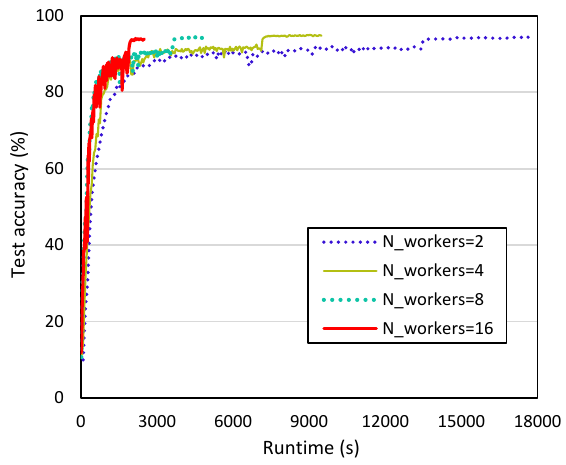}
        \caption{ResNet-152 on CIFAR-10 ($d=0.001$)}
        \label{fig:8a}
    \end{subfigure}
    ~ 
    \begin{subfigure}[t]{0.324\textwidth}
        \centering
        \includegraphics[width=1.0\linewidth]{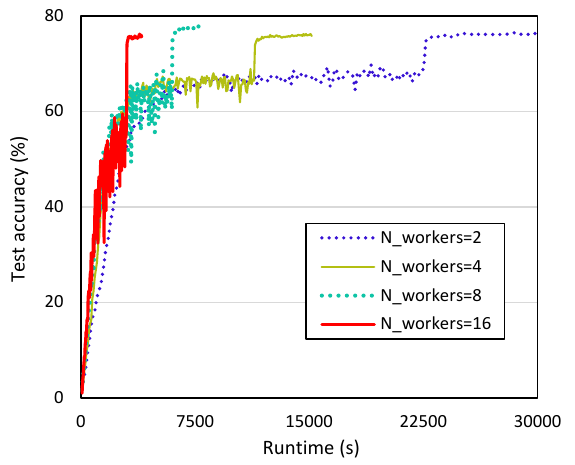}
        \caption{Inception-v4 on CIFAR-100 ($d=0.001$)}
        \label{fig:8b}
    \end{subfigure}
    ~ 
    \begin{subfigure}[t]{0.319\textwidth}
        \centering
        \includegraphics[width=1.0\linewidth]{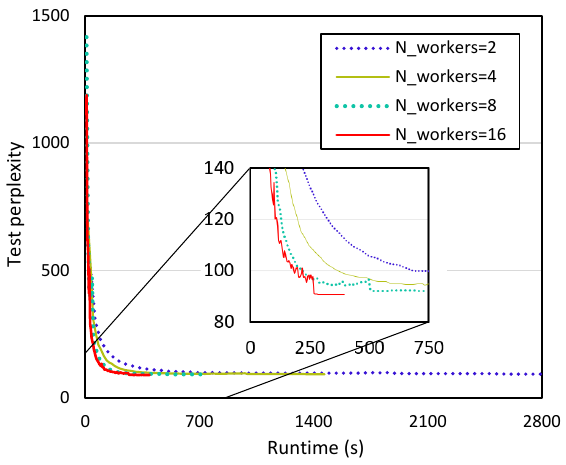}
        \caption{LSTM on WikiText-2 ($d=0.001$)}
        \label{fig:8c}
    \end{subfigure}
    \caption{Convergence performance of ExDyna by scale-out.}
    \label{fig:8}
\end{figure*}

\begin{figure*}[t]
    \centering
    \begin{subfigure}[t]{0.315\textwidth}
        \centering
        \includegraphics[width=1.0\linewidth]{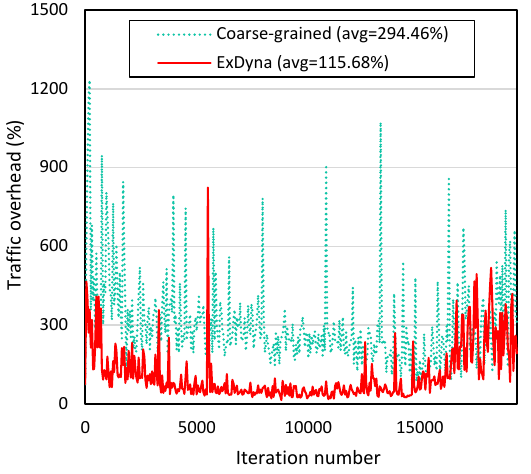}
        \caption{ResNet-152 on CIFAR-10 ($d=0.001$)}
        \label{fig:9a}
    \end{subfigure}
    ~ 
    \begin{subfigure}[t]{0.315\textwidth}
        \centering
        \includegraphics[width=1.0\linewidth]{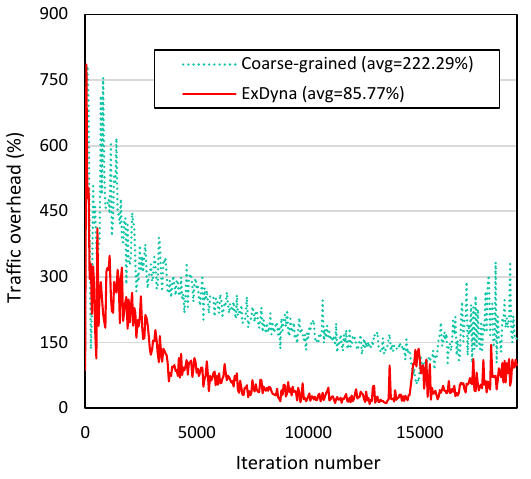}
        \caption{Inception-v4 on CIFAR-100 ($d=0.001$)}
        \label{fig:9b}
    \end{subfigure}
    ~ 
    \begin{subfigure}[t]{0.334\textwidth}
        \centering
        \includegraphics[width=1.0\linewidth]{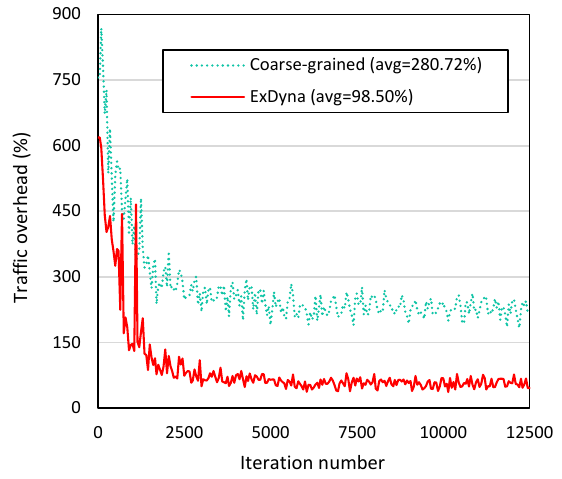}
        \caption{LSTM on WikiText-2 ($d=0.001$)}
        \label{fig:9c}
    \end{subfigure}
    \caption{Ratio of communication traffic increased by all-gather in percentage. All experiments were conducted on 16 GPUs.}
    \label{fig:9}
\end{figure*}

\begin{figure*}[t]
    \centering
    \begin{subfigure}[t]{0.314\textwidth}
        \centering
        \includegraphics[width=1.0\linewidth]{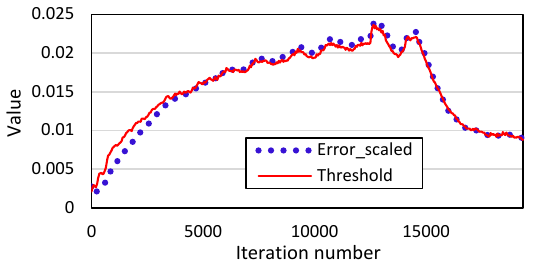}
        \caption{ResNet-152 on CIFAR-10 ($d=0.001$)}
        \label{fig:10a}
    \end{subfigure}
    ~ 
    \begin{subfigure}[t]{0.315\textwidth}
        \centering
        \includegraphics[width=1.0\linewidth]{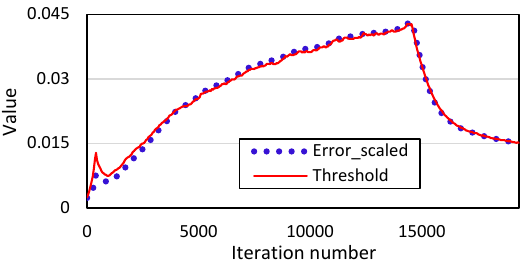}
        \caption{Inception-v4 on CIFAR-100 ($d=0.001$)}
        \label{fig:10b}
    \end{subfigure}
    ~ 
    \begin{subfigure}[t]{0.328\textwidth}
        \centering
        \includegraphics[width=1.0\linewidth]{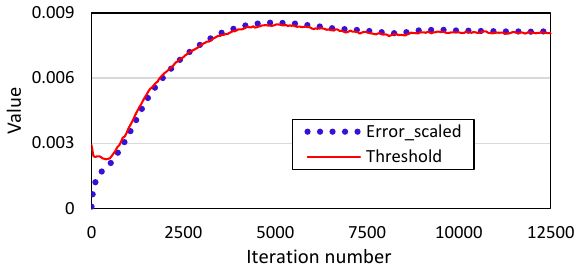}
        \caption{LSTM on WikiText-2 ($d=0.001$)}
        \label{fig:10c}
    \end{subfigure}
    \caption{Threshold estimation performance of ExDyna on 16 GPUs.}
    \label{fig:10}
\end{figure*}

\subsection{Efficiency evaluation}\label{sec:5.3}
\textbf{Scalability}. Figure~\ref{fig:8} shows the convergence performance of ExDyna by scale-out. In every experiment, ExDyna shows that the convergence performance is consistently achieved regardless of scale-out. This scalability of ExDyna mainly results from minimizing the communication overhead caused by gradient build-up, inaccurate threshold, and workload imbalance.

\textbf{Gradient vector partitioning performance}. Figure~\ref{fig:9} shows the ratio of communication traffic increased by all-gather operation in percentage. This ratio is calculated using $f(t)$ in (5). In every experiment, ExDyna significantly reduced the communication traffic overhead based on dynamic allocation of block-based partitions. That is, ExDyna dynamically adjusts the sizes of partitions to balance the workload imbalance of gradient selection between workers. Owing to the balanced workload, zero-padding overhead of the all-gather operation is significantly reduced in ExDyna. Meanwhile, the coarse-grained partitioning shows comparably imbalanced workload owing to the static topology of partitions.

\textbf{Threshold estimation performance}. To satisfy the user-required communication traffic, a sparsifier's threshold should trace the variation of the global error. Otherwise, the purpose of sparsified distributed training cannot be accomplished owing to unpredictable high actual density, like with the hard-threshold sparsifier in Figure~\ref{fig:6}. To identify whether the online threshold scaling of ExDyna can accomplish this purpose, we compared the trends of threshold and global error by scaling the magnitude of the global error. For scaling, the global error was multiplied by $\sum_{j=0}^{T-1}{\delta_{j}}/\sum_{j=0}^{T-1}{{\lVert}e_{j}{\rVert}}$, where $T$ is the number of iterations.

Figure~\ref{fig:10} shows the threshold estimation performance of ExDyna. In every experiment, the threshold of ExDyna properly traced the trend of the global error. Therefore, ExDyna can satisfy the user-set density owing to the online threshold scaling, along with the block-based partitioning, which eliminates gradient build-up.

\section{Conclusion}\label{sec:6}
In this paper, we proposed ExDyna, which divides the gradient vector into block-based partitions and allocates these partitions to workers. To alleviate the workload imbalance of gradient selection between workers, ExDyna dynamically adjusts the topology of partitions. Based on the partition allocation, each worker exclusively selects gradients in its allocated partition, which reduces computational cost through parallelization. Moreover, ExDyna estimates the accurate threshold that satisfies the user-set density using the online threshold scaling without additional overhead. These processes of ExDyna achieve high performance in terms of convergence, sparsification, gradient vector partitioning, and threshold estimation. Accordingly, ExDyna can enhance the scalability of distributed training systems remarkably. Consequently, ExDyna outperformed state-of-the-art gradient sparsifiers in our evaluation on DNN applications.

\section*{Acknowledgment}
The authors would like to thank the anonymous reviewers for their insightful feedback. This work was partially supported by RS-2023-00216370, RS-2023-00255968, and RS-2023-00283799 funded by Korean government.

\end{document}